\documentclass[journal]{IEEEtran}

\usepackage{xcolor,soul,framed} 

\colorlet{shadecolor}{yellow}
\usepackage[pdftex]{graphicx}
\graphicspath{{../pdf/}{../jpeg/}}
\DeclareGraphicsExtensions{.pdf,.jpeg,.png}

\usepackage[cmex10]{amsmath}
\usepackage{array}
\usepackage{mdwmath}
\usepackage{mdwtab}
\usepackage{eqparbox}
\usepackage{url}

\usepackage{amssymb}
\usepackage{multirow}
\newcommand{\RNum}[1]{\uppercase\expandafter{\romannumeral #1\relax}}

\begin{document}
\bstctlcite{IEEEexample:BSTcontrol}
    \title{Continuous Sign Language Recognition via Temporal Super-Resolution Network}
  \author{Qidan~Zhu,
      Jing~Li,
      Fei~Yuan,
      Quan~Gan

  \thanks{This work was supported in part by the Development Project of Ship Situational Intelligent Awareness System, China under Grant MC-201920-X01, in part by the National Natural Science Foundation of China under Grant 61673129. (Corresponding author: Jing Li)}
  \thanks{The authors are with the Key laboratory of Intelligent Technology and Application of Marine Equipment, Ministry of Education, College of Intelligent Systems Science and Engineering, Harbin Engineering University, Harbin, 150001, China (email: zhuqidan@hrbeu.edu.cn, ljing@hrbeu.edu.cn, bohelion@hrbeu.edu.cn, gquan@hrbeu.edu.cn)}
  }

\markboth{IEEE TRANSACTIONS ON MICROWAVE THEORY AND TECHNIQUES, VOL.~60, NO.~12, DECEMBER~2012
}{Roberg \MakeLowercase{\textit{et al.}}: High-Efficiency Diode and Transistor Rectifiers}

\maketitle

\begin{abstract}
Aiming at the problem that the spatial-temporal hierarchical continuous sign language recognition model based on deep learning has a large amount of computation, which limits the real-time application of the model, this paper proposes a temporal super-resolution network(TSRNet). The data is reconstructed into a dense feature sequence to reduce the overall model computation while keeping the final recognition accuracy loss to a minimum. The continuous sign language recognition model(CSLR) via TSRNet mainly consists of three parts: frame-level feature extraction, time-series feature extraction and TSRNet, where TSRNet is located between frame-level feature extraction and time-series feature extraction, which mainly includes two branches: detail descriptor and rough descriptor. The sparse frame-level features are fused through the features obtained by the two designed branches as the reconstructed dense frame-level feature sequence, and the  connectionist temporal classification(CTC) loss is used for training and optimization after the time-series feature extraction part. To better recover semantic-level information, the overall model is trained with the self-generating adversarial training method proposed in this paper to reduce the model error rate. The training method regards the TSRNet as the generator, and the frame-level processing part and the temporal processing part as the discriminator. In addition, in order to unify the evaluation criteria of model accuracy loss under different benchmarks, this paper proposes word error rate deviation(WERD), which takes the error rate between the estimated word error rate (WER) and the reference WER obtained by the reconstructed frame-level feature sequence and the complete original frame-level feature sequence as the WERD. Experiments on two large-scale sign language datasets demonstrate the effectiveness of the proposed model.
\end{abstract}


%
\IEEEpeerreviewmaketitle


\section{Introduction}

\IEEEPARstart{S}{ign} language is a basic communication tool between normal people and hearing-impaired people or between hearing-impaired people\cite{wei2020semantic}. The transmission of sign language information includes not only gestures and hand shapes, but also facial expressions and body postures. Sign language also has its own vocabulary like normal language. In the process of using sign language to communicate, vocabulary information is conveyed among people through one or a group of gesture actions as a bridge of information\cite{rastgoo2021sign}\cite{elakkiya2021machine}.\par

Video-based sign language recognition was used to identify isolated words in sign language in the early days. A video clip corresponds to a sign language word without considering the continuity of sign language. In order to perform sign language recognition more accurately, the researchers conducted a continuous sign language recognition (CSLR) study, the purpose of which was to convert a sign language video into a continuous sign language vocabulary. Due to the huge cost of frame-level annotation when creating continuous sign language video datasets, continuous sign language datasets often only have video-level annotations but no frame-level annotations, which is why researchers usually regard CSLR as weakly supervised learning\cite{koller2017re}. In view of the fact that the dataset has no frame-level annotations in CSLR, many deep learning models have been proposed and applied in CSLR\cite{han2022sign}\cite{khedkar2021analysis}\cite{adaloglou2021comprehensive}\cite{wadhawan2021sign}. According to the methods of spatiotemporal feature extraction in these models, we divide the deep learning-based models for CSLR into two classes: One is the spatial-temporal hierarchical model, which first extracts feature information from frame-level images, then extracts temporal feature information based on continuous frame-level feature sequences, and finally identification and classification are performed. The other type is the non-spatial-temporal hierarchical model, which directly extracts spatial and temporal feature information from the video for identification and classification.\par

\begin{figure*}
  \begin{center}
  \includegraphics[width=3.5in]{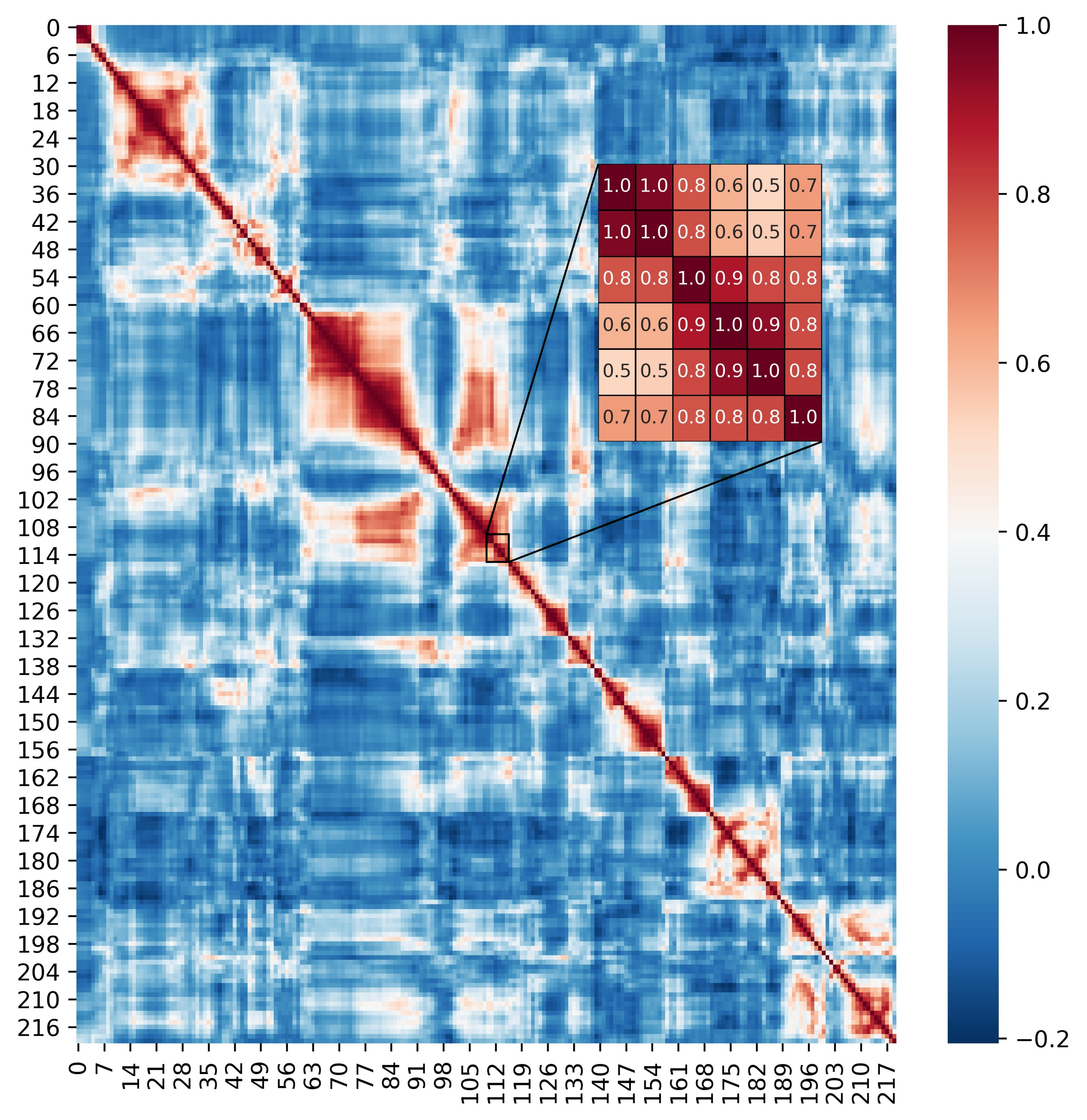}\\
  \caption{Example of Autocorrelation Matrix Visualization Plot for Frame-Level Feature Sequences of Video Data.}\label{fig:ljxy1}
  \end{center}
\end{figure*}

In the spatial-temporal hierarchical model, the extraction of spatial and temporal information from continuous sign language data is performed in series. Usually, convolutional neural network(CNN)\cite{krizhevsky2012imagenet} is used for spatial embedding, high-dimensional feature information is extracted from low-dimensional image information, and then recurrent neural network(RNN)\cite{cui2019deep}, Transformer\cite{de2020sign} or Long Short Term Memory(LSTM)\cite{tran2020overall} is used for processing in the temporal dimension to obtain high-dimensional sparse semantic information for final recognition and classification. The spatial-temporal hierarchical model has the characteristics of simple model, few parameters and clear layers, etc. It is the mainstream direction of model research in CSLR at present. However, because the spatial and temporal information are extracted and fused separately, there will inevitably be a loss of spatial and temporal information in this process, which will affect the final recognition accuracy. In the non-spatial-temporal hierarchical model, the spatial and temporal information extraction of data is carried out in parallel, that is, the data is processed in the spatial and temporal dimensions at the same time, usually using 3D-CNN\cite{li2020spatio}\cite{huang2018attention}, 2+1D-CNN\cite{koishybay2021continuous}, spatial-temporal Transformer\cite{liu2021st} and other methods. The non-spatial-temporal hierarchical model can obtain more spatial-temporal information, but its model parameters are large and the model is cumbersome. Since reducing the model calculation amount will inevitably lead to the loss of model accuracy, this paper aims to reduce the model calculation amount and improve the real-time performance under a certain range of accuracy loss. This paper mainly focuses on the deep learning-based spatial-temporal hierarchical CSLR model. In our research on the spatial-temporal hierarchical model, we found that: 1) The images between adjacent frames have high similarity in content, as shown in Figure 1, after feature extraction of frame-level images, adjacent feature vectors in the generated autocorrelation matrix have high similarity. 2) In the spatial-temporal hierarchical model, most of the computation of the model is concentrated on the extraction of video frame-level features, and the feature information extraction for each frame of the video is independent of each other. Therefore, this paper reduces the computational complexity of the model by reconstructing the sparse data into dense feature sequences while keeping the original spatial-temporal hierarchical model unchanged.\par

In this paper, a temporal super-resolution network(TSRNet) is proposed to reconstruct the sparse feature sequence into a dense feature sequence, which mainly includes two branches: a detailed descriptor and a rough descriptor, and the two types of features obtained from the two branches are fused as the reconstructed features. The model for CSLR based on this network is on the basis of the spatial-temporal hierarchical CSLR model MSTNet\cite{zhu2022multi} proposed in our previous research. First, the source video data is sparsely sampled to reduce the computational load of the frame-level feature extraction part in the spatial-temporal hierarchical model, and then the dense frame-level feature sequence reconstruction is performed through the TSRNet, and then the CTC loss\cite{graves2006connectionist} is used for training optimization after the time-series feature extraction part. The overall model is trained by the self-generating adversarial method proposed in this paper. The TSRNet is regarded as the generator, and the frame-level processing part and the time series processing part are regarded as the discriminator. The training process is divided into two steps. In addition, this paper also proposes WERD as a new criterion for evaluating the effectiveness of the proposed network. The error rate between the estimated WER and the reference WER obtained respectively by the reconstructed frame-level feature sequence and the complete original frame-level feature sequence is taken as WERD.\par

The main three contributions of this paper are as follows:\par

\begin{itemize}
\item[$\bullet$] A TSRNet is proposed, which greatly reduces the computational complexity of the original spatial-temporal hierarchical CSLR model, and the network can be flexibly inserted into any spatial-temporal hierarchical CSLR model.
\item[$\bullet$] The WERD is proposed as a new criterion to unify the criterion for model accuracy loss under different benchmarks.
\item[$\bullet$] A self-generating adversarial training method is proposed to reduce the final error.
\end{itemize}

\section{Related work}

This section will review related research from two aspects: existing CSLR methods, and related video super-resolution methods.\par

\subsection {Continuous Sign Language Recognition}
Video-based CSLR is the translation of continuous sign language videos into understandable written phrases or spoken words. In the early days of CSLR, methods such as Hidden Markov Model(HMM) were usually used for recognition, but with the development of deep learning, CNNs were introduced into CSLR, and combined with HMM to form a "CNN+HMM" hybrid model\cite{koller2017re}\cite{koller2016deep}. Koller et al.\cite{koller2018deep} embedded a CNN into an HMM end-to-end while interpreting the CNN output in a Bayesian framework, and the hybrid CNN-HMM combines the strong discriminative ability of CNNs with the sequence modeling ability of HMMs. Camgoz et al.\cite{koller2019weakly} follow a hybrid approach to embed a powerful CNN-LSTM model into each HMM stream to discover properties that themselves lack sufficient discriminative power for recognition. With the emergence of RNN and CTC, the hybrid model composed of "CNN+RNN+CTC" using RNN instead of HMM is widely used in continuous sign language recognition\cite{wei2020semantic}\cite{al2021deep}. The hybrid model mainly uses 2D-CNN for frame-level feature extraction, then uses RNN for time series processing, and finally uses CTC for training and decoding. Huang et al.\cite{huang2021boundary} developed a novel boundary-adaptive encoder-based approach for sign language recognition combined with window attention, achieving competitive results on popular benchmarks. Gao et al.\cite{gao2021rnn} proposed an efficient RNN converter-based approach for Chinese sign language processing, and designed a multi-level visual-level transcription network with frame-level, lexical-level and phrase-level BiLstm to explore multi-scale visual-semantic features . Min et al.\cite{min2021visual} proposed visual alignment constraints to enable CSLR networks to be end-to-end trainable by enforcing feature extractors to predict with more alignment supervision to address the overfitting problem of CTC in sign language recognition. These methods can be classified as spatial-temporal hierarchical models and are the most widely used CSLR methods.\par

In addition, some non-spatial-temporal hierarchical models such as "3D-CNN" and "2+1D-CNN" are also used in CSLR. Although the hybrid model of "CNN+RNN+CTC" can effectively recognize continuous sign language, the extraction of spatial-temporal features is separated. In order to extract spatial-temporal features more effectively, the 3D-CNN method is applied in CSLR\cite{sharma2021asl}\cite{han2022efficient}. Ariesta et al.\cite{ariesta2018sentence} proposed a sentence-level sign language method for deep learning combining 3D-CNN and Bi-RNN. Specifically, a 3D-CNN is used to extract features from each video frame, a Bi-RNN is used to extract unique features from the sequential behavior of video frames, and then a possible sentence is generated. Han et al.\cite{han2022sign} used "2+1D-CNN" for feature extraction and proposed a lightweight spatiotemporal channel attention module, including two sub-modules channel temporal attention and spatiotemporal attention, by combining squeeze and excitation attention combined with self-attention enables the network to focus on important information in spatial, temporal, and channel dimensions.\par

This paper mainly conducts research on the basis of the spatial-temporal hierarchical model. First, frame-level feature extraction is performed, then feature reconstruction is performed through a TSRNet, and time-series feature extraction is performed. Finally, CTC loss is used for identification and classification.\par

\subsection {Video Super Resolution}
Video super-resolution is an extension of the image super-resolution task, which restores low-resolution video images into high-resolution video images. It can fully utilize the inter-frame information in the restoration process to obtain better performance. Video super-resolution can be divided into two classes according to whether adjacent frames are aligned with the target frame: alignment methods and non-alignment methods. Alignment methods include motion compensation methods and deformable convolution methods, and non-alignment methods include spatial non-alignment methods and spatial-temporal non-alignment methods\cite{liu2022video}\cite{song2021multi}\cite{song2022learning}\cite{zhu2021video}. These methods are all aimed at restoring the spatial resolution of the video. In video super-resolution, there is another method aimed at restoring the video timing resolution, which is video frame interpolation. This method is different from the method of restoring the video from low resolution to high resolution, in that it needs to make full use of the inter-frame information to restore the current frame in the time dimension. Chen et al.\cite{cheng2021multiple} proposed an enhanced deformable separable network for video frame interpolation by processing adaptive offsets, kernels, masks, and biases learned from information in non-local neighborhoods, which has fewer parameters and improves the performance of kernel-based methods. Kalluri et al.\cite{kalluri2020flavr} propose a fully end-to-end trainable flowless method for multi-frame video interpolation, which aims to implicitly infer nonlinear motion trajectories and complex occlusions from unlabeled videos, and greatly simplifies the process of training, testing, and deploying frame interpolation models. In this study, our proposed method is most similar to the video frame interpolation method, which utilizes the inter-frame feature information to recover adjacent feature information in the temporal dimension using frame interpolation.\par

\begin{figure*}
  \begin{center}
  \includegraphics[width=5in]{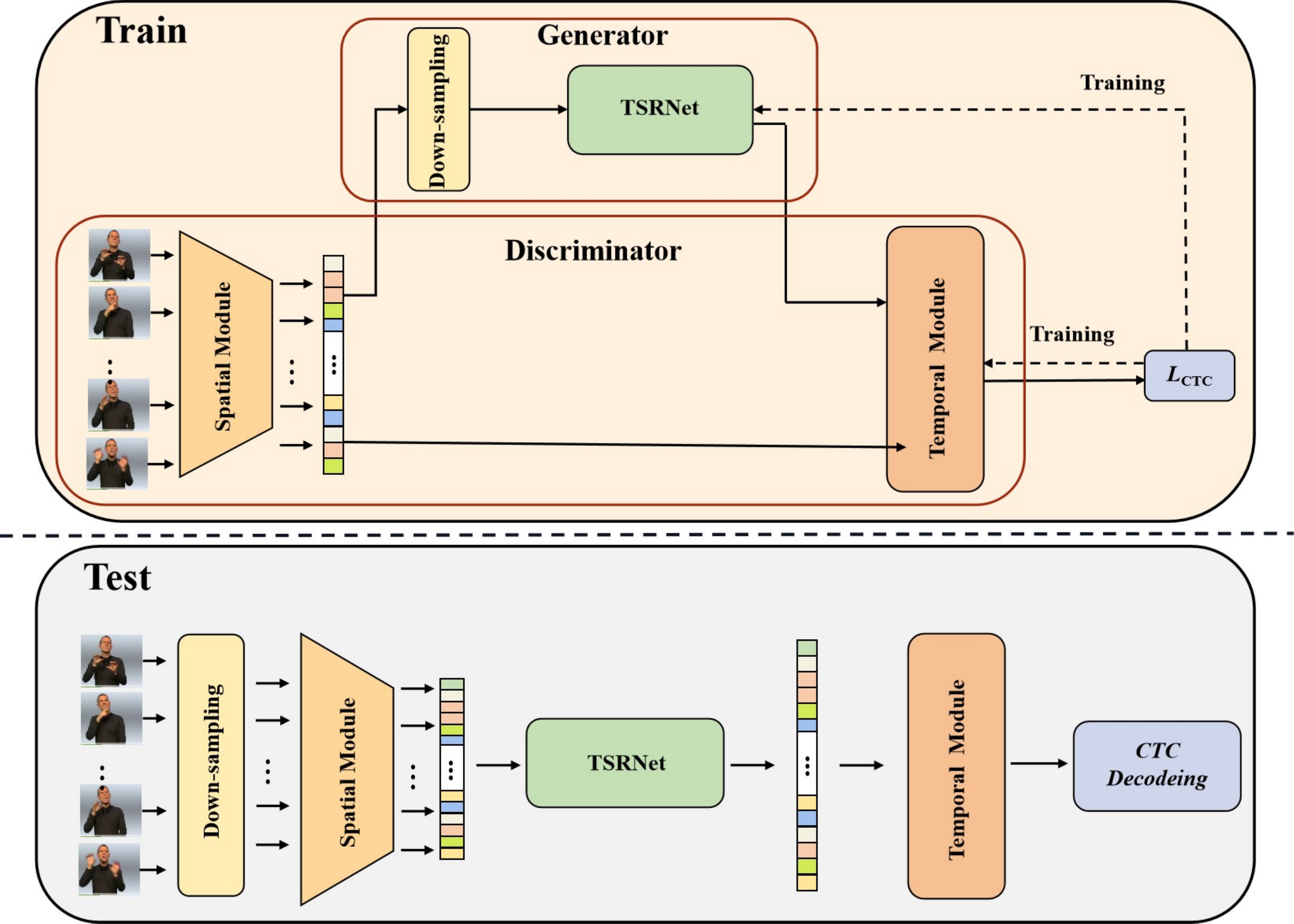}\\
  \caption{Overall architecture diagram of continuous sign language recognition model via temporal super-resolution network.}\label{fig:ljxy2}
  \end{center}
\end{figure*}

\section{Methodology}
The overall architecture of the CSLR model via TSRNet proposed in this paper is shown in Figure 2. The model mainly consists of three parts: frame-level feature extraction, time series processing and TSRNet. In the model training stage, for the input sign language video, the frame-level feature sequence is first obtained through the frame-level feature extraction part. After down-sampling, a dense frame-level feature sequence is obtained through the TSRNet. Then, the final time series features are obtained through the time series processing part. Finally, the CTC loss is used for training optimization, and the sign language recognition results are obtained. The entire training phase is trained using our proposed self-generating adversarial training method, where the temporal super-resolution network is regarded as the generator, and the frame-level processing part and the temporal processing part are regarded as the discriminator. The network is trained using the down-sampled data of the frame-level feature sequence as the input of the temporal super-resolution network. And the training is divided into two steps, the first step is to train the spatial-temporal hierarchical model, and the second step is to train the TSRNet. In the model testing stage, the position of down-sampling is different from that in the training stage. At this time, the input sign language video is directly down-sampled and then the frame-level feature sequence is obtained through the frame-level feature extraction part. Then a dense frame-level feature sequence is obtained through the temporal super-resolution network TSRNet. Then go through the time series processing part. For the obtained timing features, use CTC to decode to get the final sign language recognition result.\par

The CSLR model via temporal super-resolution network is based on our previous research MSTNet\cite{zhu2022multi}. The frame-level feature extraction part and the time-series feature extraction part of the model are consistent with MSTNet, that is, the frame-level features are based on resnet-34, and the time-series features are extracted using the "1DCNN+Transformer" coding structure. In order to further reduce the computational cost of the model and improve the real-time performance of the model, this paper proposes the TSRNet. The details of the TSRNet will be introduced in section A. The proposed specific training and testing methods will be introduced in detail in section C.\par

\subsection {Temporal Super-Resolution Network}

The temporal super-resolution network mainly includes a detail descriptor and a coarse descriptor, as shown in Figure 3. The main branch is the detail descriptor. For the frame-level feature sequence with unified channel dimensions, the detailed description of the feature sequence is obtained by convolution of multiple 1-D residual blocks and 1-D transpose. Another branch is the rough descriptor, which directly up-samples the frame-level feature sequence to obtain a rough description of the feature sequence. The input of the temporal super-resolution network is the sparse frame-level feature sequence extracted by the spatiotemporal hierarchical model, and the output is the feature sequence obtained by fusing the detailed features and rough features. The feature sequence is the reconstructed dense frame-level feature sequence, which is used as the input of the sequential processing part of the subsequent spatiotemporal hierarchical model.\par

\begin{figure*}
  \begin{center}
  \includegraphics[width=5in]{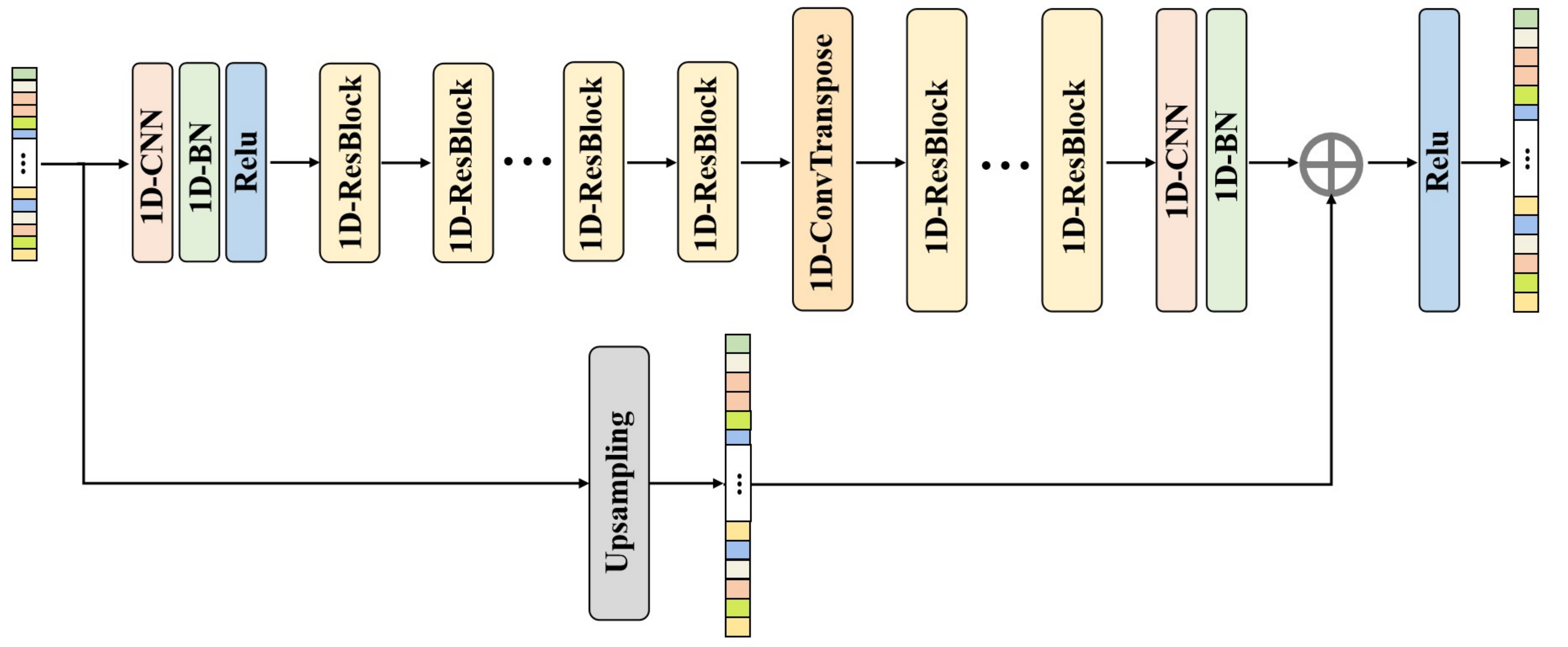}\\
  \caption{Framework of Temporal Super-Resolution Network.}\label{fig:ljxy3}
  \end{center}
\end{figure*}

For an input sign language video $V=(x_1,x_2,...,x_T)=\{{x_t|_1^T\in \mathbb{R}^{T\times c\times h\times w}}\}$ containing $T$ frame, where $x_t$ is the t-th frame image in the video, $h\times w$ is the size of $x_t$, and $c$ is the number of channels. $V$ passes through the frame-level feature extractor $F_s$ of the spatial-temporal hierarchical model, and obtains the feature expression as follows:\par

\begin{equation}
f_1 = F_s(V)\in \mathbb{R}^{T\times c_1}
\end{equation}

where $c_1$ is the number of channels after feature extraction.\par

The dense frame-level feature sequence $f_1$ is down-sampled by $n$ times to obtain a sparse frame-level feature sequence, and then the time dimension and the channel dimension are exchanged to obtain the feature sequence $f_2\in \mathbb{R}^{c_1\times T_1}$, where $T_1=T/n$.\par

At this time, $f_2$ is the sparse feature sequence, which is the input of the temporal super-resolution network. Input $f_2$ to the two branches of the temporal super-resolution network, namely the detail descriptor and the coarse descriptor, respectively.\par

In the detail descriptor branch, the number of channels of the sparse feature sequence is first dimensionally increased by a 1D-CNN, and then batch normalized and activated by the activation function to obtain the sparse feature sequence $f_3\in \mathbb{R}^{c_2\times T_1}$. The activation function $\sigma$ is Relu, and $c_2$ is the number of channels after the ascension. Then the dimension-raising process can be described as:\par

\begin{equation}
f_3 = \sigma(BN(1D-CNN(f_2)))\in \mathbb{R}^{c_2\times T_1}
\end{equation}

Assuming that the entire dimension-raising process is $F_{1DCNN-Relu}$, equation(2) can be described as:\par

\begin{equation}
f_3 = F_{1DCNN-Relu}(f_2)\in \mathbb{R}^{c_2\times T_1}
\end{equation}

Then, for the sparse feature sequence $f_3$, the detailed features are extracted through several 1D-Resblocks to obtain the sparse feature sequence $f_3^{'}$. The number of channels and the time length of $f_3^{'}$ remain unchanged, which is consistent with $f_3$.\par

\begin{equation}
f_3^{'} = F_{Res_m}(f_3)\in \mathbb{R}^{c_2\times T_1}
\end{equation}

where $F_{Res_m}$ indicates that the number of 1D-Resblocks is $m$.\par

After that, use 1-dimensional transposed convolution for the sparse feature sequence $f_3^{'}$ in the time dimension, and up-sampling $n$ times to obtain the dense feature sequence $f_4\in \mathbb{R}^{c_3\times T}$, and the corresponding number of channels is also increased in dimension, $c_3$ is the number of channels after the up-dimension, $T$ is the original video frame length. For $f_4$, the process of formula (4) is performed again, and the dense feature sequence $f_4^{'}=F_{Res_k}(f_4)\in \mathbb{R}^{c_3\times T}$ is obtained by extracting features through $k$ 1D-Resblocks, and the number of channels and time length of $f_4^{'}$ remain unchanged.\par

$f_4^{'}$ is subjected to a 1D-CNN for dimensionality reduction, so that the number of channels is restored to be consistent with the input feature sequence, and then batch normalization is performed to obtain a dense feature sequence $f_5$.\par

\begin{equation}
f_5 = BN(1D-CNN(f_4^{'}))\in \mathbb{R}^{c_1\times T}
\end{equation}

Assuming that the entire dimensionality reduction process is $F_{1DCNN}$, equation (5) can be described as:\par

\begin{equation}
f_5 = F_{1DCNN}(f_4^{'})\in \mathbb{R}^{c_1\times T}
\end{equation}

In the rough descriptor branch, only the nearest neighbor interpolation $F_{nearest}$ is used for the sparse feature sequence $f_2$ to up-sample n times in the time dimension, resulting in a dense feature sequence $f_2^{'}\in \mathbb{R}^{c_1\times T}$\par

\begin{equation}
f_2^{'} =F_{nearest}(f_2)\in \mathbb{R}^{c_1\times T}
\end{equation}

Finally, the dense feature sequences $f_2^{'}$ and $f_5$ obtained by the two branches are fused, and the final output dense feature sequence   is obtained through the activation function. $f_{Result}$ is the dense frame-level feature sequence reconstructed by the temporal super-resolution network.\par

\begin{equation}
f_{Result} =\sigma(f_2^{'}+f_5)
\end{equation}

Resblock, an important component in our proposed TSRNet, was originally proposed by He et al.\cite{he2016deep} in 2016, and its purpose is to address the problem of network degradation. That is, as the number of layers of the network deepens, the accuracy of the network decreases instead. The proposal of Resblock effectively addresses this problem, which makes the number of layers of the network increase sharply. Initially, ResBlock was used in classification tasks. Due to its excellent performance, it was introduced into other related fields and achieved excellent results. In the proposed model ConvNeXt, Liu et al.\cite{liu2022convnet} verified that the replacement of ordinary convolution with depth-wise convolution in ResBlock has no effect on model performance, but the amount of parameters is greatly reduced. This paper introduces the ResBlock structure into our model. Because we deal with 1D temporal data, we use 1D depth-wise convolution in 1D-Resblock, and its structure is shown in Figure 4.\par

\begin{figure}
  \begin{center}
  \includegraphics[width=1in]{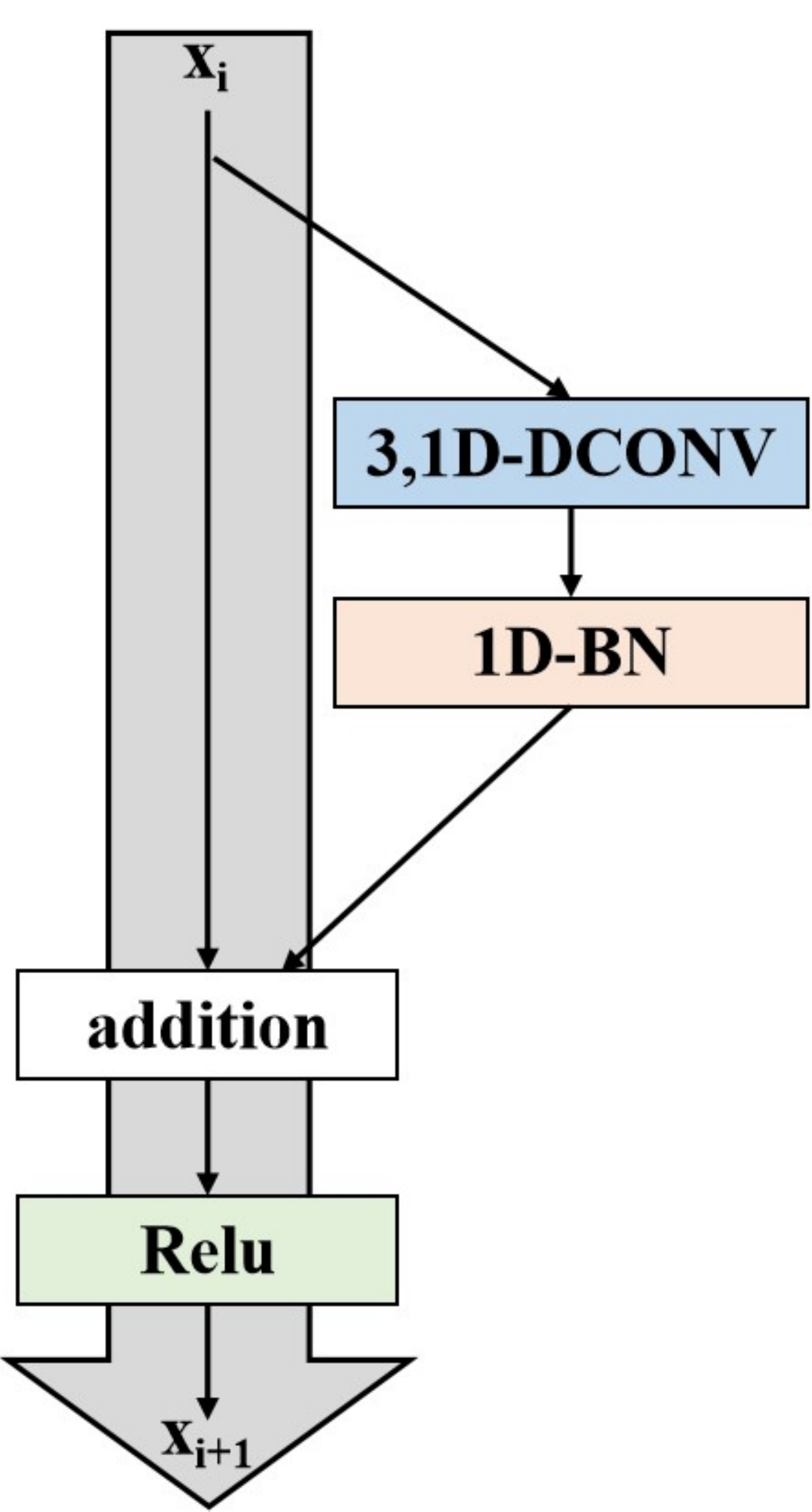}\\
  \caption{Structure of 1D-ResBlock.}\label{fig:ljxy4}
  \end{center}
\end{figure}

Let the 1D-ResBlock map of layer be:

\begin{equation}
y =\sigma(BN(F_d(x)+x)
\end{equation}

where $x$ is the input data, $y$ is the output data, and $F_d(\cdot)$ is the 1D depth-wise convolution.\par

\subsection {Connectionist Temporal Classification}

CSLR belongs to weakly supervised learning. The input video is an unsegmented sequence and lacks a strict correspondence between video frames and labeled sequences. After encoding the input video sequence, it is very appropriate to use CTC as a decoder. CTC was originally designed for speech recognition, mainly to perform end-to-end temporal classification of unsegmented data to address the problem of mismatched lengths of input and output sequences. In recent years, it is often used in CSLR. In the CSLR model via TSRNet proposed in this paper: in the model training stage, for the final dense feature sequence, CTC Loss is used to train and optimize to obtain the optimal model; in the model testing stage, for the final time series features, using CTC decoding to get the final sign language recognition result.\par

CTC introduces a blank label $\{-\}$ to mark unclassified labels during decoding, that is, any words in the input video clip that do not belong to the sign language vocabulary, so that the input and output sequences can be matched, and the dynamic programming method is used for decoding\cite{min2021visual}.\par

For the input video $V$ of $T$ frames, the label of each frame is represented by $\pi=(\pi_1,\pi_2,...,\pi_T)$, where $\pi_T\in v\cup \{-\}$, and $v$ is the sign language vocabulary, then the posterior probability of the label is:\par

\begin{equation}
p(\pi|V)=\prod_{\substack{t=1}}^{\substack{T}}p(\pi_t|V)=\prod_{\substack{t=1}}^{\substack{T}}Y_{t,\pi_t}
\end{equation}

For a given sentence-level label $s=(s_1,s_2,...,s_L)$, where L is the number of words in the sentence. CTC defines a many-to-one mapping B, whose operation is to remove blank labels and duplicate labels (eg, $B(-g-re-e-n-)=B(-gr-e-en-)=green$) in the path, then the conditional probability of label s is the sum of the occurrence probabilities of all corresponding paths:\par

\begin{equation}
p(s|V)=\sum_{\substack{\pi\in B^{-1}(s)}}p(\pi|V)
\end{equation}

Where $B^{-1}(s)=\{\pi|B(\pi)=s\}$ is the inverse mapping of B. CTC loss is defined as the negative log-likelihood of the conditional probability of s.\par

\begin{equation}
L_{CTC}=-\ln p(s|V)
\end{equation}

The final sign language recognition result is obtained by CTC decoding after normalization by Softmax function.\par

\begin{figure}
  \begin{center}
  \includegraphics[width=2.5in]{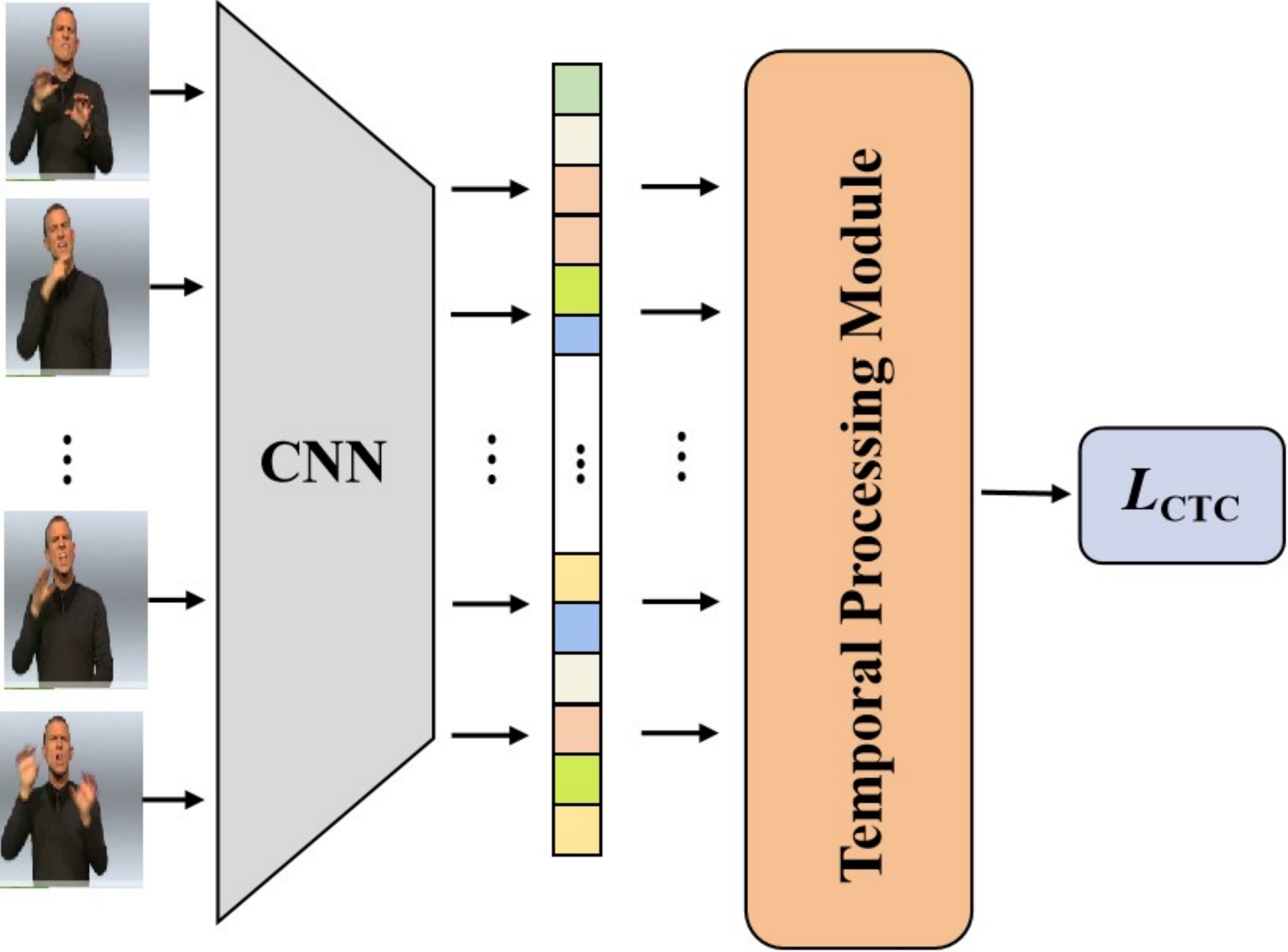}\\
  \caption{The spatial-temporal hierarchical model architecture.}\label{fig:ljxy5}
  \end{center}
\end{figure}

\subsection {Model Training and Testing Process}

The TSRNet proposed in this paper belongs to the super-resolution model. The commonly used training method in super-resolution models is to use L1 loss or L2 loss as the loss function, which reflects the accuracy of the estimated value by judging the distance between the estimated value and the reference value, considering the gap of a single data level\cite{dong2015image}\cite{wang2019end}. However, what this paper wants to reconstruct is the feature vector. If only the gap between the data is considered, the result is bound to be unsatisfactory. Because each feature vector is a whole, it represents the integrated high-dimensional information, and the value may be very small after multiple feature extractions, so the similarity between vectors needs to be considered at this time. This paper draws on the training method of generative adversarial network(GAN)\cite{radford2015unsupervised}, applies it to our model training, and proposes a self-generating adversarial training method to train the temporal super-resolution network, which greatly improves the final error rate. The training and testing process of the self-generating adversarial training method is described in detail below.\par

\emph{Training:} During the training process, we use the self-generating adversarial training method to train the temporal super-resolution network. We consider the temporal super-resolution network as the generator and the spatial-temporal hierarchical model as the discriminator. First, the original sign language video is input into the frame-level feature extraction part of the spatial-temporal hierarchical model to obtain the frame-level feature sequence, and the down-sampling data is used as the input of the temporal super-resolution network to train the network. And the training is divided into two steps, the first step is to train the spatial-temporal hierarchical model, and the second step is to train the TSRNet.\par

\begin{figure*}
  \begin{center}
  \includegraphics[width=5in]{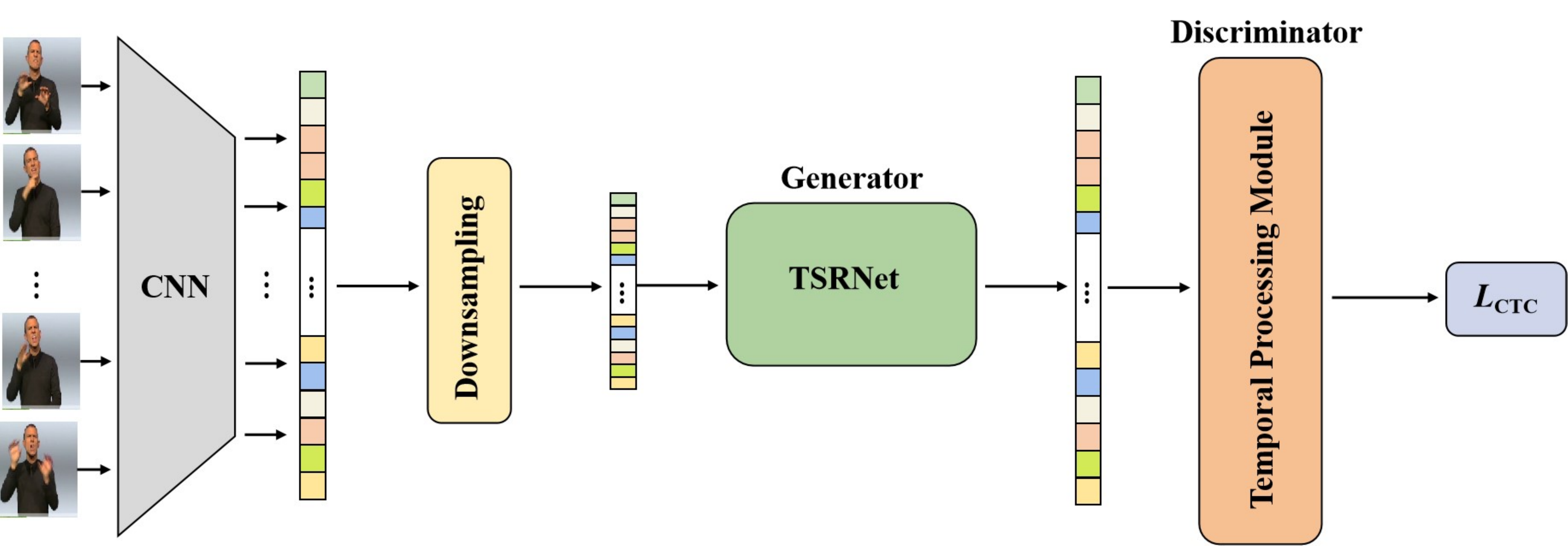}\\
  \caption{Overall model architecture after inserting temporal super-resolution network.}\label{fig:ljxy6}
  \end{center}
\end{figure*}

Step 1: Train the spatial-temporal hierarchical model, as shown in Figure 5. The original sign language video data is used as the input of CNN to extract frame-level features, and then the processed time-series features are obtained through the time-series processing module. Finally, the labeled video-level phrases are used as labels, and CTC Loss is used for training, and the final model obtained is used as the discriminator.\par

Step 2: Train the TSRNet, as shown in Figure 6. First, insert the TSRNet between the frame-level feature extraction part and the time-series feature extraction part of the spatial-temporal hierarchical model, freeze the parameters of the spatial-temporal hierarchical model trained in the first step, and only train the TSRNet parameters. Then, using the original sign language video as input, the frame-level feature sequence is obtained through the frame-level feature extraction part of the spatial-temporal hierarchical model, and the frame-level feature sequence is sparse according to the set down-sampling multiple. Then the sparse frame-level feature sequence is input into the temporal super-resolution network for reconstruction, and the reconstructed dense frame-level feature sequence is obtained. Finally, the sequence is input into the time-series processing part of the spatial-temporal hierarchical model to obtain the final time-series features, which are trained using CTC loss according to the phrase annotation.\par

Here, in order to increase the robustness, proportional random sampling is used in the process of sparseness. Assuming that the multiple of down-sampling is 4 and the length of the frame-level feature sequence is T, then the dense frame-level feature sequence slices the sequence with a width of 4 and divides it into $n$ segments, where $n=T/4$. Then a random feature vector is taken from each segment and reconstructed into a sparse feature sequence, which is the input of the TSRNet.\par

\emph{Testing:} During testing, the difference from training is the location of down-sampling. Firstly, the input video data is directly down-sampled to obtain sparse video data, which is input to the frame-level extraction part of the spatial-temporal hierarchical model to obtain frame-level features. Then, a dense frame-level feature sequence is obtained by reconstruction through the TSRNet. The resulting dense frame-level feature sequence is then input into the temporal processing part of the spatial-temporal hierarchical model. Finally, for the obtained timing features, the final recognition result is obtained through CTC decoding, as shown in the test part of Figure 2.\par

\section{Experiment}

In this section, we conduct comprehensive experiments on two widely used sign language recognition datasets to verify the effectiveness of the proposed model. A series of ablation experiments are performed to demonstrate the effect of each component of the proposed model. For the evaluation criteria proposed in this paper, we describe in detail in section C.\par

\subsection {Dataset}

RWTH-PHOENIX-Weather-2014(RWTH) dataset\cite{koller2015continuous}: RWTH is recorded by a public weather radio station in Germany. All presenters wear dark clothing and performed sign language in front of a clean background. The videos in this dataset are recorded by 9 different presenters with a total of 6841 different signed sentences (where the number of sign language word instances is 77321 and the number of vocabulary words is 1232). All videos are preprocessed to a resolution of $210\times 260$, and a frame rate of 25 frames per second (FP/S). The dataset is officially divided into 5,672 training samples, 540 validation samples, and 629 test samples.\par

Chinese Sign Language(CSL) dataset\cite{huang2018video}: CSL is captured using a Microsoft Kinect camera and contains 100 sentences of everyday Chinese language, each sentence demonstrates 5 times by 50 presenters with a vocabulary size of 178. The video resolution is $1280\times 720$ and the frame rate is 30 FP/S. The performance diversity of this dataset is richer because the demonstrators wear different clothes and demonstrate different speeds and ranges of motion. In the absence of official segmentation, we divide the CSL dataset into a training set and a test set according to the 8:2 rule. The training set accounts for 80\% and the test set accounts for 20\%, that is, it is divided into a training set of 20,000 samples and a test set of 5,000 samples, and make sure that the sentences in the training and test sets are the same, but the presenters are different.\par

\begin{table*}[!htbp]
\centering
\caption{Comparison of the experimental results of TSRNet on the RWTH dataset and the other two methods}
\label{tab:aStrangeTable1}
\begin{tabular}{c|c|c|c|c|c|c|c|c|c|c|c|c}
\hline  
\multirow{3}{*}{Down-sampling factor}& \multicolumn{4}{c|}{TSRNet}& \multicolumn{4}{c|}{Nearest neighbor interpolation}& \multicolumn{4}{c}{Linear interpolation}\\
\cline{2-13}
 & \multicolumn{2}{c|}{WER(\%)}& \multicolumn{2}{c|}{WERD(\%)}& \multicolumn{2}{c|}{WER(\%)}& \multicolumn{2}{c|}{WERD(\%)}& \multicolumn{2}{c|}{WER(\%)}& \multicolumn{2}{c}{WERD(\%)}\\
\cline{2-13}
  & dev& test& dev& test& dev& test& dev& test& dev& test& dev& test\\
 \hline  
 1& 20.3& 21.4& 0& 0& 20.3& 21.4& 0& 0& 20.3& 21.4& 0& 0\\
 \hline  
 2& 20.7& 21.5& 1.9& 0.5& 21.8& 22.3& 7.0& 4.3& 22.9& 23.4& 12.3& 9.5\\
 \hline  
 3& 21.1& 22.2& 3.8& 3.8& 24.8& 25.5& 21.1& 19.3& 26.2& 26.4& 27.4& 23.4\\
 \hline  
 4& 23.4& 24.7& 14.7& 15.6& 26.3& 27.4& 27.8& 27.8& 29.5& 30.3& 41.2& 40.0\\
 \hline  
 5& 25.4& 25.3& 23.8& 18.4& 30.5& 30.2& 45.1& 39.6& 33.9& 33.5& 57.0& 52.0\\
 \hline  
 6& 28.2& 28.9& 36.0& 34.3& 33.5& 33.9& 55.7& 53.4& 38.3& 38.4& 69.5& 67.0\\
 \hline  
 7& 31.1& 31.5& 47.4& 44.7& 38.5& 38.9& 70.0& 68.3& 44.4& 44.0& 81.7& 79.2\\
 \hline  
 8& 35.3& 34.9& 61.4& 56.7& 43.8& 43.4& 80.8& 78.1& 51.1& 50.3& 89.9.4& 88.0\\
 \hline  
\end{tabular}
\end{table*}

\subsection {Implementation Rules}

In the overall model of this paper, the Adam optimizer\cite{kingma2014adam} is used for training, the initial learning rate and weight factor are set to $10^{-4}$, and the batch size used is 2. During model training, random cropping and random flipping are used for data augmentation. For random cropping, the frame size of each video sequence is first resized to $256\times 256$, and then randomly cropped to a size of $224\times 224$ to fit the shape of the input. For random flips, set flip probability to 0.5. Flip and crop processing is performed on video sequences. In addition, temporal enhancement processing is performed to randomly increase or shorten the length of the video sequence within $\pm 20\%$. During model testing, only center cropping is used for data enhancement, and the beam search algorithm is used for decoding in the final CTC decoding stage, with a beam width of 10. For the RWTH dataset, there are 30 epochs in the training phase, and the learning rate is reduced by 80\% at the 10th and 20th epochs. For the CSL dataset, there are 15 epochs in the training phase, and the learning rate is reduced by 80\% at the 5th and 10th epochs. The graphics card used in this experiment is RTX2080Ti, the GPU dedicated memory size is 12G, the CPU memory is 8G, and the number of cores is 4.\par

\subsection {Judgment Criteria}

The WER is widely used as a criterion for CSLR\cite{koller2015continuous}. It is the sum of the minimum insertion operations, replacement operations, and deletion operations required to convert the recognition sequence into a standard reference sequence. Lower WER means better recognition performance, which is defined as follows:\par

\begin{equation}
WER=100\%\times \frac{ins+del+sub}{sum}
\end{equation}

where “ins” represents the number of words to be inserted, “del” represents the number of words to be deleted, “sub” represents the number of words to be replaced, and “sum” represents the total number of words in the label.\par

For the evaluation of the performance of the model in this paper, only using WER can represent the recognition performance of the model, but it cannot accurately represent the gap between the recognition results of the model in this paper and the recognition results of the original model. Therefore, this paper proposes WERD to further measure the performance of this model. The frame-level feature sequence reconstructed by the model and the complete original frame-level feature sequence are both processed through the time-series processing part of the spatial-temporal hierarchical model and the final CTC decoding to obtain the final recognition result, and the WER is calculated to obtain the estimated WER and reference WER respectively, and then the error rate between them is calculated as a further evaluation criterion for the model in this paper.\par

\begin{equation}
WERD=100\%\times \frac{1-1.1^{-(WER_E-WER_R)}}{1+1.1^{-(WER_E-WER_R)}}
\end{equation}

where $WER_E$ represents the estimated $WER$ and $WER_R$ represents the reference WER. Because $WER_E$ is a process that approximates $WER_R$, $WER_E$ is greater than or equal to $WER_R$. The smaller the $WERD$, the closer the estimated value is to the reference value, which means that the dense feature sequence reconstructed by the model is more similar to the original feature sequence. Ideally, $WERD=0$, that is, the estimated sequence is equal to the original sequence. When experimenting on the CSL dataset, we treat a single Chinese character as a word.\par

\subsection {Experimental Results}

In this paper, the model MSTNet in our previous study is used as the base model, and the TSRNet is inserted on its basis. The original data of the RWTH dataset and CSL dataset are down-sampled by different multiples as the input of the overall model, and the feature sequence is reconstructed using the proposed TSRNet. The experimental results are presented in Table \RNum{1} and Table \RNum{2}, respectively. The experimental results are compared with the other two methods for data recovery after down-sampling, and the comparison results are also presented in Table \RNum{1} and Table \RNum{2}. To further analyze the experimental data, the experimental results in Tables \RNum{1} and \RNum{2} are presented as graphs, as shown in Figures 7 and 8.\par

\begin{table*}[!htbp]
\centering
\caption{Comparison of the experimental results of TSRNet on the CSL dataset and the other two methods}
\label{tab:aStrangeTable2}
\begin{tabular}{c|c|c|c|c|c|c}
\hline  
\multirow{2}{*}{Down-sampling factor}& \multicolumn{2}{c|}{TSRNet}& \multicolumn{2}{c|}{Nearest neighbor interpolation}& \multicolumn{2}{c}{Linear interpolation}\\
\cline{2-7}
 & WER(\%)& WERD(\%)& WER(\%)& WERD(\%)& WER(\%)& WERD(\%)\\
\hline
 1& 0.7& 0& 0.7& 0& 0.7& 0\\
 \hline  
 2& 0.7& 0& 0.7& 0& 0.7& 0\\
 \hline  
 3& 0.7& 0& 0.8& 0.5& 0.8& 0.5\\
 \hline  
 4& 0.7& 0& 0.9& 1.0& 0.9& 1.0\\
 \hline  
 5& 0.7& 0& 0.9& 1.0& 1.0& 1.4\\
 \hline  
 6& 0.8& 0.5& 1.3& 2.9& 1.4& 3.3\\
 \hline  
 7& 0.9& 1.0& 1.8& 5.2& 2.1& 6.7\\
 \hline  
 8& 1.1& 1.9& 2.6& 9.0& 3.1& 11.4\\
 \hline  
 9& 1.5& 3.8& 3.6& 13.7& 4.6& 18.4\\
 \hline  
 10& 1.9& 5.7& 5.4& 22.0& 6.9& 28.7\\
 \hline  
 11& 2.5& 8.6& 7.9& 32.6& 10.0& 41.6\\
 \hline  
 12& 3.8& 14.7& 11.1& 45.9& 13.6& 54.7\\
 \hline  
 13& 5.1& 20.7& 14.4& 57.4& 17.0& 65.1\\
 \hline  
 14& 6.9& 28.7& 17.6& 66.7& 21.3& 75.4\\
 \hline  
 15& 8.4& 35.1& 20.4& 73.5& 24.0& 80.3\\
 \hline  
 16& 10.3& 42.8& 24.0& 80.3& 28.3& 86.6\\
 \hline  
\end{tabular}
\end{table*}


\begin{figure*}
  \begin{minipage}{0.5\textwidth}
\includegraphics[width=3.5in,height=2.33in]{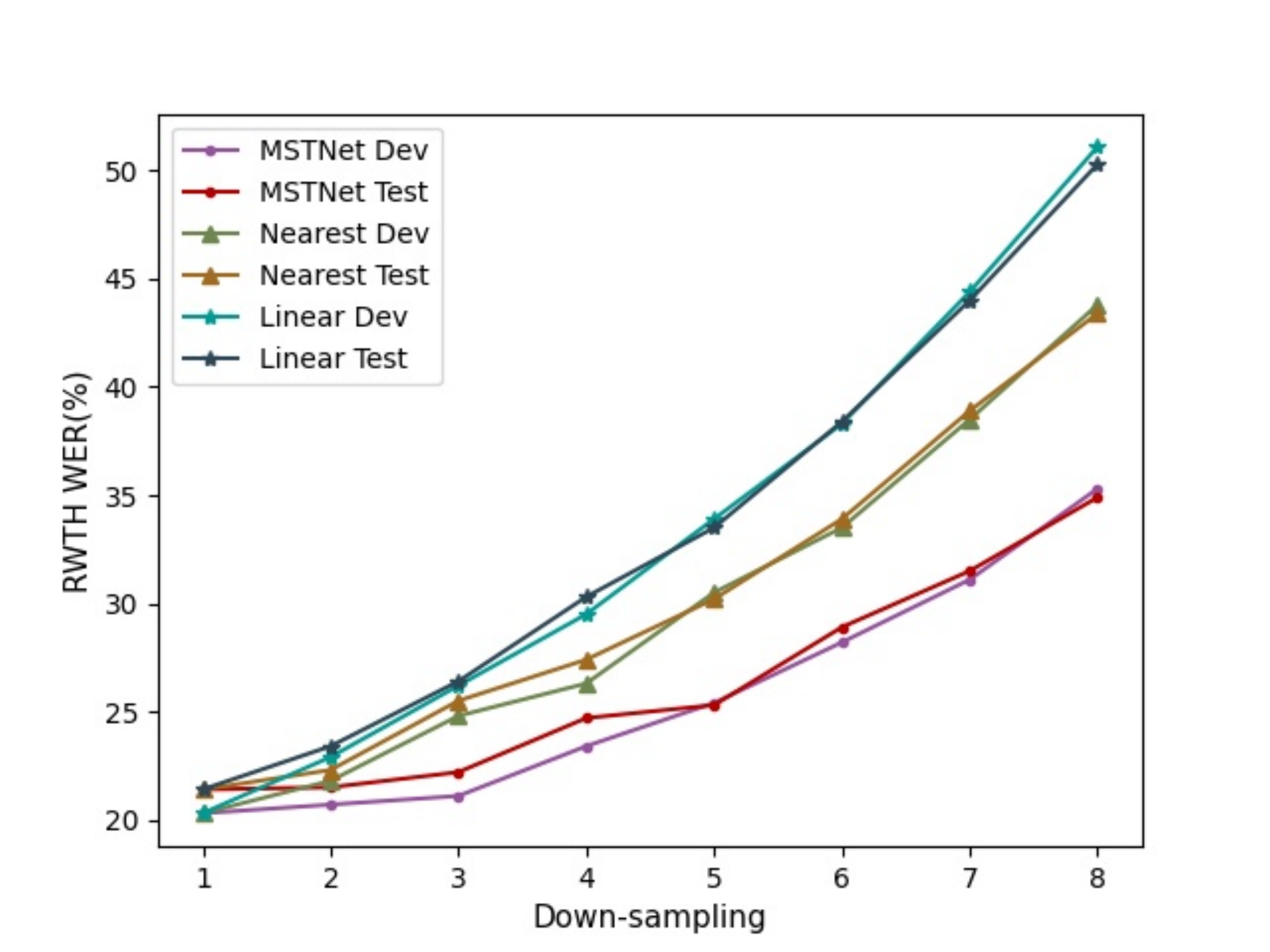}
\caption{Comparison of the WER of TSRNet on the RWTH dataset and the other two methods.}
\label{fig:ljxy7}
\end{minipage}
\hfill
\begin{minipage}{0.5\textwidth}
\includegraphics[width=3.5in,height=2.33in]{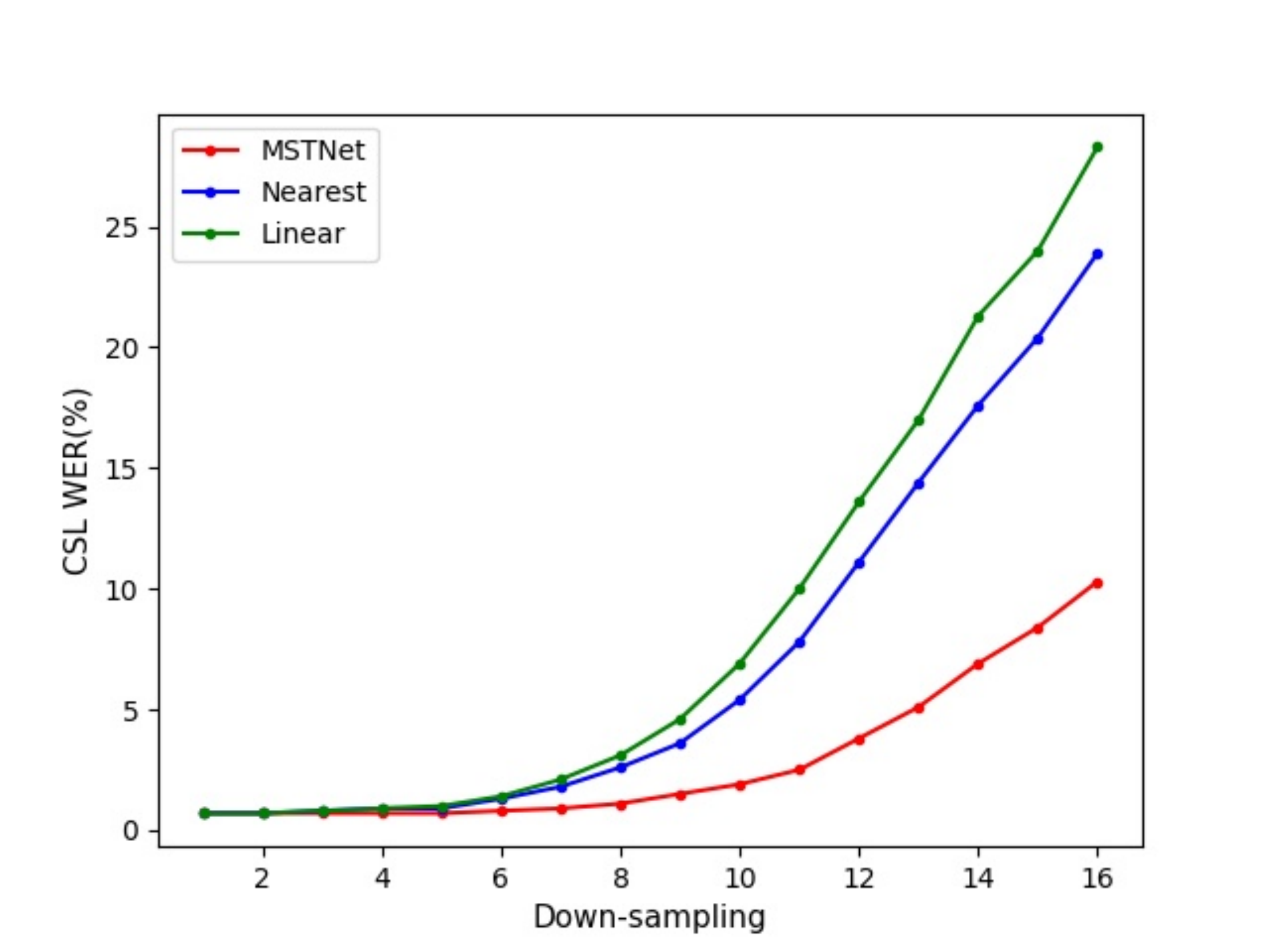}
\caption{Comparison of the WER of TSRNet on the CSL dataset and the other two methods.}
\label{fig:ljxy8}
\end{minipage}
\end{figure*}

From Table \RNum{1} and Table \RNum{2}, it can be obtained: Using different methods to reconstruct the down-sampled data under the same down-sampling factor, the final WERD obtained by our proposed temporal super-resolution network is smaller than the nearest neighbor interpolation method and the linear interpolation method, and when the down-sampling factor is larger, the advantage achieved by our method is more obvious.\par

For the RWTH dataset, the reference WER obtained on the validation set and test set, that is, the WER obtained without down-sampling the data, is 20.3\% and 21.4\%, respectively. When the down-sampling factor is 2, the WERD obtained by our method on the validation set and the test set is 1.9\% and 0.5\%, respectively, and the WERD using the nearest neighbor interpolation method is 7.0\% and 4.3\%, compared with our method, the error increased by 5.1\% and 3.8\% respectively, the WERD using the linear interpolation method is 12.3\% and 9.5\%, and the error is increased by 10.4\% and 9\% respectively compared with our method. When the down-sampling factor is 3, the errors on the validation set and test set using the nearest neighbor interpolation method are increased by 17.3\% and 15.5\%, respectively. Compared with our method, the errors obtained by using the linear interpolation method are increased by 23.6\% and 19.6\% respectively. When the down-sampling factor is 4, the errors on the validation set and the test set using the nearest neighbor interpolation method are increased by 13.1\% and 12.2\% respectively compared with our method, and compared with our method, the errors obtained by using the linear interpolation method are respectively increased by 26.5\% and 24.4\%. By analogy, the error is increasing.\par

\begin{table*}[!htbp]
\centering
\caption{Overall performance of continuous sign language recognition model via temporal super-resolution network}
\label{tab:aStrangeTable3}
\begin{tabular}{c|c|c|c|c|c|c|c|c}
\hline  
Down-sampling factor& 1& 2& 3& 4& 5& 6& 7& 8\\
\hline
\multicolumn{9}{c}{Frame-level feature extraction part}\\
\hline
Params& $23\times 10^6$& $23\times 10^6$& $23\times 10^6$& $23\times 10^6$& $23\times 10^6$& $23\times 10^6$& $23\times 10^6$& $23\times 10^6$\\
\hline  
FLOPs(GFlops)& 734.74& 367.37& 242.47& 183.69& 149.95& 121.23& 102.87& 91.84\\
\hline  
MemR+W(GByte)& 14.94& 7.47& 4.93& 3.73& 2.99& 2.46& 2.09& 1.87\\
\hline  
\multicolumn{9}{c}{Temporal Super-Resolution Network}\\
\hline  
Params& -& $61\times 10^6$& $61\times 10^6$& $61\times 10^6$& $61\times 10^6$& $61\times 10^6$& $61\times 10^6$& $61\times 10^6$\\
\hline  
FLOPs(GFlops)& -& 11.57& 11.41& 11.25& 11.19& 11.37& 11.28& 11.09\\
\hline  
MemR+W(GByte)& -& 0.36& 0.35& 0.34& 0.34& 0.34& 0.34& 0.33\\
\hline  
\multicolumn{9}{c}{Time-series feature extraction part}\\
\hline  
Params& $97\times 10^6$& $97\times 10^6$& $97\times 10^6$& $97\times 10^6$& $97\times 10^6$& $97\times 10^6$& $97\times 10^6$& $97\times 10^6$\\
\hline  
FLOPs(GFlops)& 1.79& 1.79& 1.79& 1.79& 1.79& 1.79& 1.79& 1.79\\
\hline  
MemR+W(GByte)& 0.035& 0.035& 0.035& 0.035& 0.035& 0.035& 0.035& 0.035\\
\hline  
\multicolumn{9}{c}{Total}\\
\hline  
Params& $120\times 10^6$& $181\times 10^6$& $181\times 10^6$& $181\times 10^6$& $181\times 10^6$& $181\times 10^6$& $181\times 10^6$& $181\times 10^6$\\
\hline  
FLOPs(GFlops)& 736.53& 380.73& 253.88& 196.73& 159.93& 134.39& 115.94& 104.72\\
\hline  
MemR+W(GByte)& 14.94& 7.87& 5.31& 4.11& 3.36& 2.83& 2.46& 2.24\\
\hline  
Params MEM(MB)& 458.75& 691.63& 691.63& 691.63& 691.63& 691.63& 691.63& 691.63\\
\hline  
Run Time(ms)& 221.57& 137.71& 100.78& 73.5& 63.93& 58.74& 52.08& 48.56\\
 \hline  
\end{tabular}
\end{table*}


\begin{figure*}
  \begin{minipage}{0.5\textwidth}
\includegraphics[width=3.5in,height=2.33in]{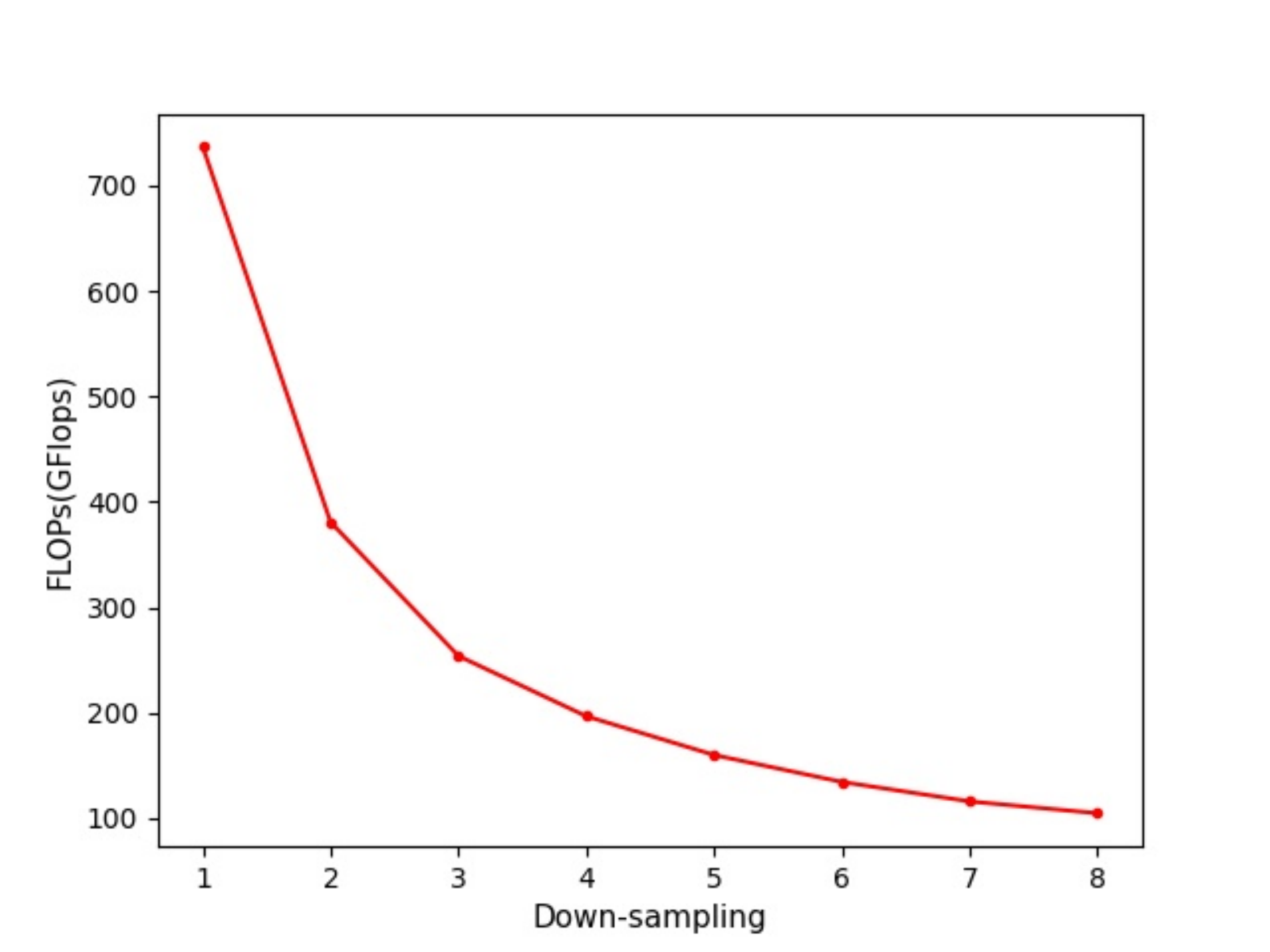}
\caption{The relationship between the down-sampling multiple and the overall model calculation amount.}
\label{fig:ljxy9}
\end{minipage}
\hfill
\begin{minipage}{0.5\textwidth}
\includegraphics[width=3.5in,height=2.33in]{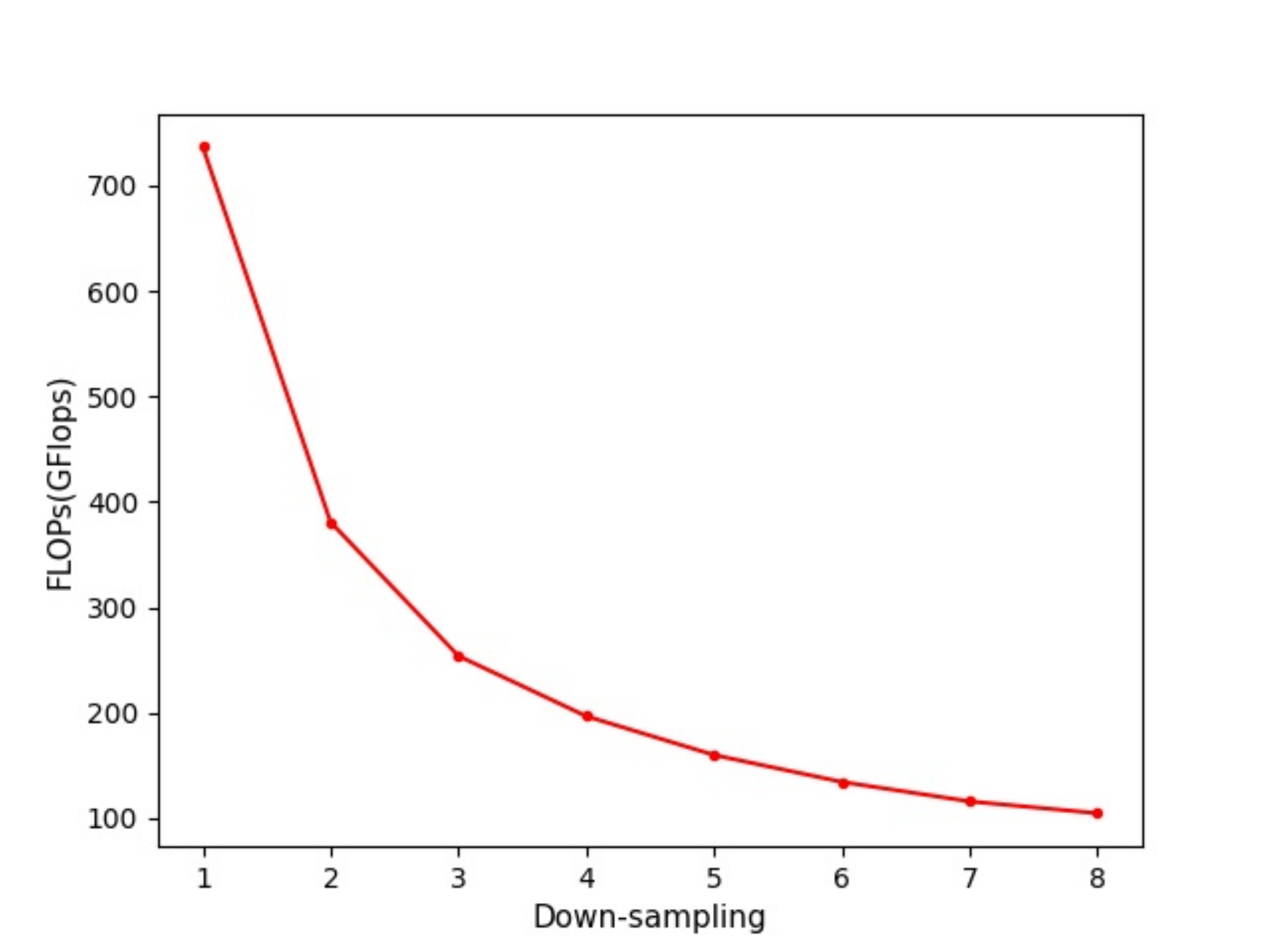}
\caption{The relationship between the down-sampling factor and the overall model running time.}
\label{fig:ljxy10}
\end{minipage}
\end{figure*}

For the CSL dataset, the reference WER for the test set is 0.7\%. When the down-sampling multiples are 2, 3, 4, and 5, the WERD obtained by our method on the test set is all 0, that is, there is no error. This is because the frame rate of CSL is 30FPS/s, and the sign language demonstrator is slow to demonstrate sign language, resulting in more redundancy in the continuous sign language dataset. When the down-sampling factor is 6, the WERD obtained by our method on the test set is 0.5\%, and the errors obtained by the nearest neighbor interpolation method and the linear interpolation method are increased by 2.4\% and 2.8\% respectively compared with our method. When the down-sampling factor is 7, the errors obtained by using the nearest neighbor interpolation method and the linear interpolation method are increased by 4.2\% and 5.7\%, respectively, compared with our method. By analogy, the error is increasing.\par

As can be seen from the curves in Figures 7 and 8, as the down-sampling factor increases, the WER obtained on the two datasets using the three methods also increases. And the WER obtained using our method TSRNet is much smaller than the nearest neighbor interpolation method and the linear interpolation method, proving the effectiveness of the TSRNet.\par

As can be seen from Figure 7, for the RWTH dataset, the WER curve of TSRNet increases abruptly after the down-sampling multiple is 3, which indicates that the down-sampling factor of 3 is an inflection point. From the calculation of the experimental results in Table \RNum{1}, it can be seen that the WERD obtained on the validation set and the test set when the down-sampling factor is 2 is 1.9\% and 0.5\% higher than that when the down-sampling factor is 1; When the down-sampling factor is 3, it increases by 1.9\% and 3.3\%, respectively, compared with the down-sampling factor of 2; when the down-sampling multiple is 4, it increases by 11.4\% and 11.7\%, respectively, compared with the down-sampling multiple of 3. This is consistent with the conclusion that the down-sampling factor of 3 is an inflection point obtained in Figure 7, which proves that TSRNet has the best model performance when the down-sampling factor is 3 on the RWTH dataset.\par

As can be seen from Figure 8, for the CSL dataset, the WER curve of TSRNet increases sharply after the down-sampling multiple is 11, which indicates that the down-sampling multiple is 11 is an inflection point. From the calculation of the experimental results in Table \RNum{2}, it can be seen that when the down-sampling multiple is 10, the WERD obtained on the test set is 1.9\% larger than that when the down-sampling multiple is 9; When the down-sampling factor is 11, the WERD increases by 2.9\% compared with the down-sampling factor of 10; When the down-sampling factor is 12, the WERD increases by 6.1\% when the down-sampling factor is 11. This is consistent with the conclusion that the down-sampling factor of 11 is an inflection point obtained in Figure 8, which proves that TSRNet has the best model performance when the down-sampling factor is 11 on the CSL dataset.\par

In order to further analyze the effectiveness and superiority of TSRNet, under different down-sampling times, this paper analyzes the overall model from three aspects: calculation amount, parameter amount, and running time. The original data, that is, the video frame length of 200 and the image size of $224\times 224$, are input into the overall model. The calculation amount, parameter amount and running time of the model under different down-sampling times are shown in Table \RNum{3}. The running time is the average of five consecutive model running times.\par

\begin{table*}[!htbp]
\centering
\caption{Overall model performance comparison of inserting TSRNet into different spatiotemporal hierarchical models}
\label{tab:aStrangeTable4}
\begin{tabular}{c|c|c|c|c|c|c|c|c|c|c|c|c}
\hline  
\multirow{3}{*}{Down-sampling factor}& \multicolumn{4}{c|}{CNN+BiLSTM+CTC}& \multicolumn{4}{c|}{VAC\cite{min2021visual}}& \multicolumn{4}{c}{MSTNet\cite{zhu2022multi}}\\
\cline{2-13}
 & \multicolumn{2}{c|}{WER(\%)}& \multicolumn{2}{c|}{WERD(\%)}& \multicolumn{2}{c|}{WER(\%)}& \multicolumn{2}{c|}{WERD(\%)}& \multicolumn{2}{c|}{WER(\%)}& \multicolumn{2}{c}{WERD(\%)}\\
\cline{2-13}
 & dev& test& dev& test& dev& test& dev& test& dev& test& dev& test\\
 \hline  
 1& 26.1& 26.7& 0& 0& 21.8& 22.8& 0& 0& 20.3& 21.4& 0& 0\\
  \hline  
 4& 29.8& 30.4& 17.5& 17.5& 25.1& 26.2& 15.6& 16.1& 23.4& 24.7& 14.7& 15.6\\
 \hline  
\end{tabular}
\end{table*}

\begin{table*}[!htbp]
\centering
\caption{Overall model performance comparison using different types of 1D-ResBlock}
\label{tab:aStrangeTable5}
\begin{tabular}{c|c|c|c}
\hline  
\multirow{2}{*}{Types of 1D-ResBlock}& \multicolumn{2}{c|}{WER(\%)}& \multirow{2}{*}{The amount of parameters of TSRNet}\\
\cline{2-3}
 & dev& test& \\
 \hline 
 A& 23.7& 24.9& $350\times 10^6$\\
 \hline  
 B& 23.4& 24.9& $230\times 10^6$\\
 \hline  
 C& 23.4& 24.7& $61\times 10^6$\\
 \hline  
 D& 23.8& 24.7& $61\times 10^6$\\
 \hline  
\end{tabular}
\end{table*}

As can be seen from Table \RNum{3}, for the amount of parameters, the introduction of TSRNet on the basis of MSTNet results in an increase of $61\times 10^6$ in the amount of parameters, and with the increase of the down-sampling multiple, the size of the parameter amount remains unchanged. As for the amount of computation, with the increase of the down-sampling multiple, the amount of computation keeps decreasing. And the model calculation amount is positively correlated with the running time, as the calculation amount decreases, the running time also decreases, as shown in Figures 9 and 10.\par

As can be seen from Table \RNum{3}, for the amount of parameters, the introduction of TSRNet on the basis of MSTNet results in an increase of $61\times 10^6$ in the amount of parameters, and with the increase of the down-sampling multiple, the size of the parameter amount remains unchanged. As for the amount of computation, with the increase of the down-sampling multiple, the amount of computation keeps decreasing. And the model calculation amount is positively correlated with the running time. As the calculation amount decreases, the running time also decreases, as shown in Figures 9 and 10. It can be seen from the previous experiments that TSRNet has the best model performance when the down-sampling multiple is 3 on the RWTH dataset. For Figure 9 and Figure 10, when the down-sampling multiple is 3, the overall calculation of the model is reduced by 65.53\% and the running time was reduced by 54.52\%.\par

Through the above experiments, it can be found that for the spatial-temporal hierarchical model, the calculation amount is mainly concentrated in the frame-level feature extraction part, and most of its parameters are concentrated in the time-series feature extraction part. The introduction of TSRNet greatly reduces the computational complexity of frame-level feature extraction, but increases the overall model parameters.\par

\subsection {Ablation Experiment}
In this section, we conduct ablation experiments on the RWTH dataset to further verify the effectiveness of each component of the model. WER and WERD are used as metrics in ablation experiments, and the down-sampling factor is set to 4.\par

\emph{1) Experiment 1:} Overall model performance of inserting TSRNet into different spatial-temporal hierarchical models. The TSRNet proposed in this paper is inserted into different spatial-temporal hierarchical models as a sub-network to reduce the overall model computation and running time with an acceptable error rate. In this paper, TSRNet is inserted into 4 different spatial-temporal hierarchical models, and the WER and WERD obtained by the overall model are calculated respectively, as shown in Table \RNum{4}.The two models compared in the Table \RNum{4} "CNN+BiLSTM+CTC" and "VAC"\cite{min2021visual} are retrained by us.\par

It can be seen from Table \RNum{4} that under the same down-sampling factor, inserting TSRNet into different spatial-temporal hierarchical models obtains less variation in the WERD of the overall model. It means that TSRNet has good generalization in different spatial-temporal hierarchical models, so that the overall performance of the model remains stable.\par

\emph{2) Experiment 2:} Overall model performance using different types of 1D-ResBlock. This paper introduces 1D-ResBlock in TSRNet, but the overall model performance obtained by using different types of 1D-ResBlock is different. This paper conducts experiments using 4 different types of 1D-ResBlock respectively, and calculates the parameters of WER and TSRNet of the overall model under different types, as shown in Table \RNum{5}. The 4 different types of 1D-ResBlock used are shown in Figure 11.\par

\begin{figure}
  \begin{center}
  \includegraphics[width=3.5in]{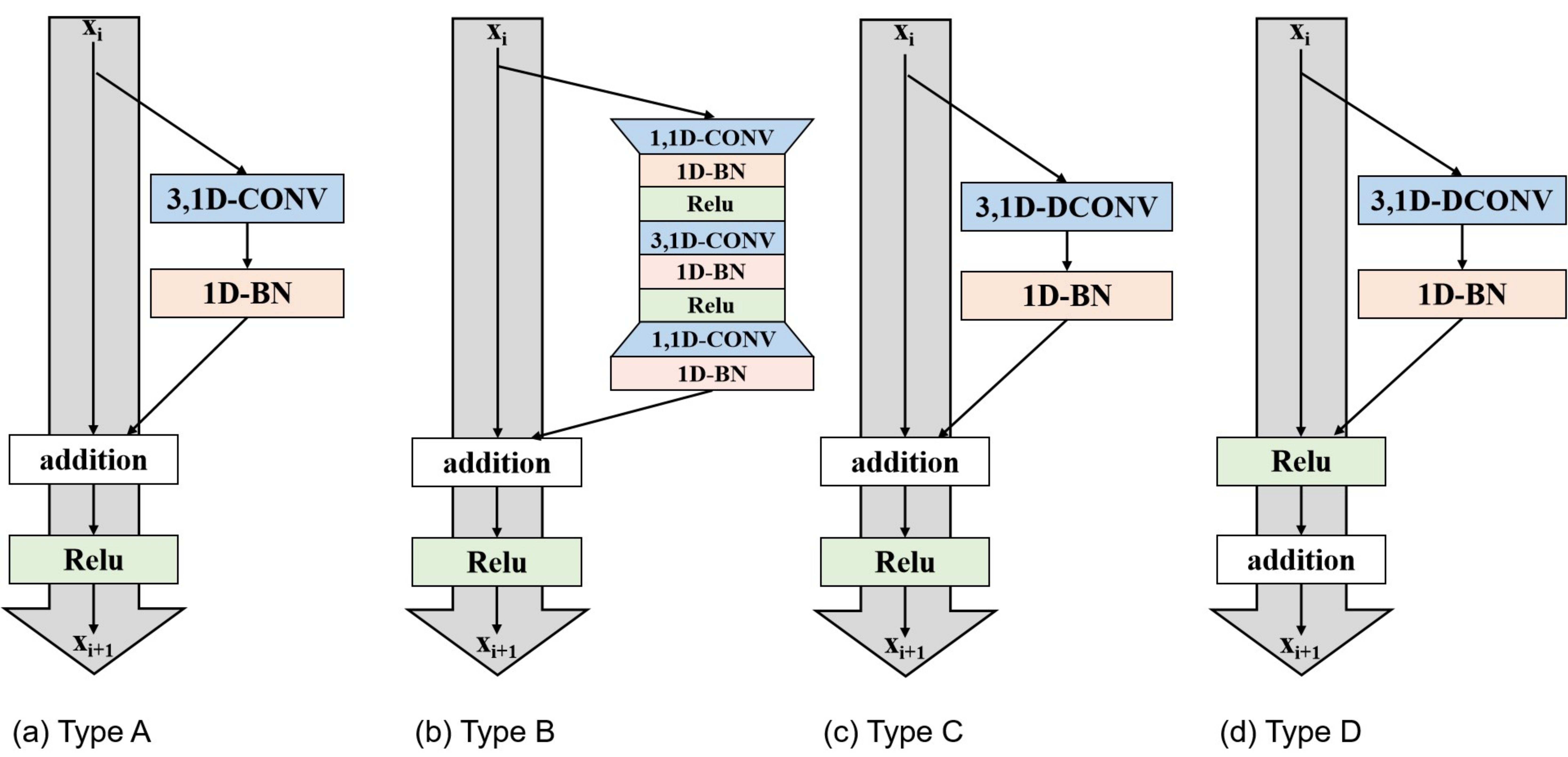}\\
  \caption{4 different types of 1D-ResBlock.}\label{fig:ljxy11}
  \end{center}
\end{figure}

It can be obtained from Table \RNum{5} that the overall model with 1D-ResBlock type C has the best performance. When 1D-ResBlock types are C and D, the number of parameters is smaller than that when types are A and B, because types C and D use depth-wise convolution 1D-DCONV. For types C and D with the same amount of parameters, type C reduces the WER by 1.7\% on the validation set compared to type D.\par

\emph{3) Experiment 3:} Overall model performance using different down-sampling methods. When down-sampling the data, this paper uses proportional random sampling. The overall model WER obtained when using different down-sampling methods is shown in Table \RNum{6}.\par

\begin{table*}[!htbp]
\centering
\caption{Overall model performance comparison using different down-sampling methods}
\label{tab:aStrangeTable6}
\begin{tabular}{c|c|c}
\hline  
\multirow{2}{*}{Down-sampling factor}& \multicolumn{2}{c}{WER(\%)}\\
\cline{2-3}
 & dev& test\\
 \hline 
 Equally spaced sampling& 23.4& 25.0\\
 \hline  
 Proportional random sampling& 23.4& 24.7\\
 \hline  
 Random sampling& 26.0& 26.5\\
 \hline  
\end{tabular}
\end{table*}

\begin{table*}[!htbp]
\centering
\caption{Overall model performance comparison using different training methods}
\label{tab:aStrangeTable7}
\begin{tabular}{c|c|c}
\hline  
\multirow{2}{*}{Training method}& \multicolumn{2}{c}{WER(\%)}\\
\cline{2-3}
 & dev& test\\
 \hline 
 Conventional super-resolution network training method& 24.9& 26.1\\
 \hline  
 Self-generating adversarial training method& 23.4& 24.7\\
 \hline  
\end{tabular}
\end{table*}

It can be seen from Table \RNum{6} that when the proportional random sampling method is used, the WER obtained by the overall model is the lowest and the performance is the best. Compared with the equally spaced sampling method, the WER obtained on the test set is reduced by 1.2\%, and the WER obtained by the random sampling method on the validation set and test set is reduced by 11.1\% and 7.3\%, respectively. Proportional sampling can ensure the integrity of video data as much as possible, and random sampling can make the overall model have better generalization.\par

\emph{4) Experiment 4:} Overall model performance using different training methods. The overall model in this paper is trained using our proposed self-generating adversarial training method. We compare the self-generating adversarial training method with the conventional super-resolution network training method, and the WER of the overall model is shown in Table \RNum{7}. The conventional super-resolution network training method here refers to: down-sample the dense feature sequence and input it into TSRNet, use L2 Loss to train the reconstructed dense feature sequence and the reference dense feature sequence, and then insert the trained model into the spatial-temporal hierarchical model to obtain the WER of the overall model.\par

As can be seen from Table \RNum{7}, the overall model WER obtained by using our proposed self-generating adversarial training method is reduced by 6.4\% and 5.7\% on the validation set and test set, respectively, compared with the conventional super-resolution network training method, proving that the effectiveness of self-generating adversarial training methods. Using conventional super-resolution network training methods only considers gaps between data levels, while ignoring gaps at semantic levels. For CSLR, semantic-level information plays an extremely important role, and the semantic-level information can be better recovered using our proposed self-generating adversarial training method.\par

\section{Conclusion}
A deep learning-based spatial-temporal hierarchical continuous sign language recognition model uses dense sampling when extracting information from raw videos. However, as the video length increases, the amount of computation increases exponentially, making the model unsuitable for processing long video data in practical applications, limiting the real-time application of the model. In response to this problem, this paper proposes a new TSRNet to reduce the computational complexity of the CSLR model and improve real-time performance. The CSLR model via TSRNet mainly consists of three parts: frame-level feature extraction, time-series feature extraction and TSRNet. The TSRNet is located in the middle of the frame-level feature extraction part and the time-series feature extraction part, and mainly includes two branches: the detail descriptor and the rough descriptor. The sparse frame-level features are fused through the features obtained by the two designed branches as the reconstructed dense frame-level feature sequence, and the CTC loss is used for training and optimization after the time-series feature extraction part. In this paper, a self-generating adversarial training method is proposed to train the model. The temporal super-resolution network is regarded as the generator, and the frame-level processing part and the time-series processing part are regarded as the discriminator, which can better restore the semantic-level information and greatly reduce the model error rate. In addition, this paper proposes WERD as a new criterion to unify the criterion for model accuracy loss under different benchmarks. Experiments on two large-scale sign language datasets demonstrate the effectiveness of the proposed model, which greatly reduces the overall model computation load and greatly improves real-time performance under a certain range of accuracy loss. And TSRNet can be flexibly inserted into any spatial-temporal hierarchical model.\par

CSLR is designed to solve the communication problem between hearing-impaired people and normal people, and the computational load and memory footprint of the model need to meet real-time requirements. The network proposed in this paper improves the real-time performance of the original model, but there are two problems, that is, it only optimizes the calculation amount of the frame-level spatial feature extraction part, and increases the memory footprint on the basis of the original model. And the model in this paper is only for the spatial-temporal hierarchical model in the CSLR model, but not involved in the non-spatial-temporal hierarchical model. Therefore, how to design a more lightweight, more real-time and more generalized model is a problem worth studying.\par

\section*{Acknowledgment}

This work was supported in part by the Development Project of Ship Situational Intelligent Awareness System, China under Grant MC-201920-X01, in part by the National Natural Science Foundation of China under Grant 61673129. \par


%





\ifCLASSOPTIONcaptionsoff
  \newpage
\fi





\bibliographystyle{IEEEtran}
\bibliography{IEEEabrv,Bibliography}

\vfill


\end{document}